\def\year{2018}\relax
\documentclass[letterpaper]{article} %DO NOT CHANGE THIS
\usepackage{aaai18}  %Required
\usepackage{times}  %Required
\usepackage{helvet}  %Required
\usepackage{courier}  %Required
\usepackage{url}  %Required
\usepackage{graphicx}  %Required
\usepackage{amsmath}
\usepackage{amssymb}
\usepackage{color}
\usepackage{booktabs}
\usepackage{multirow}
\begin{document}
	\title{Learning Binary Residual Representations for Domain-specific Video Streaming}
	\author{Yi-Hsuan Tsai$^{1}$
		\hspace{0.15in} Ming-Yu Liu$^{2}$
		\hspace{0.15in} Deqing Sun$^{2}$
		\hspace{0.15in} Ming-Hsuan Yang$^{12}$	
		\hspace{0.15in} Jan Kautz$^{2}$ \\
		\hspace{0.1in} $^{1}$University of California, Merced \hspace{0.4in} $^{2}$NVIDIA
	}
	\maketitle
	\begin{abstract}
		We study domain-specific video streaming. Specifically, we target a streaming setting where the videos to be streamed from a server to a client are all in the same domain and they have to be compressed to a small size for low-latency transmission. Several popular video streaming services, such as the video game streaming services of GeForce Now and Twitch, fall in this category. While conventional video compression standards such as H.264 are commonly used for this task, we hypothesize that one can leverage the property that the videos are all in the same domain to achieve better video quality. Based on this hypothesis, we propose a novel video compression pipeline. Specifically, we first apply H.264 to compress domain-specific videos. We then train a novel binary autoencoder to encode the leftover domain-specific residual information frame-by-frame into binary representations. These binary representations are then compressed and sent to the client together with the H.264 stream. In our experiments, we show that our pipeline yields consistent gains over standard H.264 compression across several benchmark datasets while using the same channel bandwidth.
	\end{abstract}

	\section{Introduction}
	
	Video streaming services, such as Netflix and YouTube, are popular methods of viewing entertainment content nowadays. Due to large video sizes and limited network bandwidth, video compression is required for streaming video content from a server to a client. While video compression can reduce the size of a video, it often comes with undesired compression artifacts, such as image blocking effects and blurry effects. 
	
	Decades of efforts were made towards delivering the best possible video quality under bandwidth constraint. State-of-the-art video compression methods such as MPEG-4~\cite{Li_mpeg_2001}, H.264~\cite{Wiegand_h264_2003}, and HEVC~\cite{sullivan2012overview} combine various classical techniques including image transform coding, predictive coding, source coding, and motion estimation in a carefully-engineered framework. These methods are general and can be applied to various video domains for effectively compressing most of the information in a video. However, the residual information, which is the difference between the uncompressed and compressed videos, is difficult to compress because it contains highly non-linear patterns. Neither linear predictive coding nor linear transform coding can effectively compress the residual.
	
	In this paper, we hypothesize that one can effectively compress the residual information if one is willing to limit the use of the video compression method to a specific domain. In other words, we no longer wish to have a video compression method that works universally well on all videos. We only wish to have a method that works particularly well in one specific domain. Although this setting may appear inconvenient at a first glance as one needs to have a video compressor for each domain, it does fit well with several major video streaming applications, such as video game streaming and sports streaming. For example, for video game streaming services such as GeForce Now and Twitch, the gamer chooses a game title to play, and the video content is rendered on the server and delivered to client's mobile console. During the game playing period, all the video content in the stream is in the same video game domain and a domain-specific video compression method is entirely appropriate. The setting also fits other user cases such as streaming sports, which are often limited to a particular discipline, as well as things like compressing dash cam videos, as all the videos are about street scenes.
	
	To verify our hypothesis, we leverage deep learning models, which have been established as powerful non-linear function approximators, to encode the highly nonlinear residual information. In our video compression pipeline, we first apply H.264 to compress videos in a specific domain and train a novel binary autoencoder to encode the resulting residual information frame-by-frame into a binary representation. We then apply Huffman coding~\cite{Cover_information_theory} to compress the binary representations in a lossless manner. The compressed binary representations can be sent to the client in the meta data field in the H.264 streaming packet. This way, our method can be integrated into the existing video streaming standard. We illustrate our method in Figure~\ref{fig:overview}. We show that with our proposed binary residual representation, one can achieve a better video quality (at a chosen bandwidth) by sending the video stream using H.264 at a smaller bandwidth and utilizing the saved bandwidth to transmit the learned binary residual representation.
	
	We conduct extensive experiments to verify our hypothesis that we can improve state-of-the-art video compression methods by learning to encode the domain-specific residual information in a binary form. We also compare various ways of training the proposed binary autoencoder for encoding the residual information. On the KITTI~\cite{Geiger_CVPR_2012} and three games video datasets, we show that our method consistently outperforms H.264 both quantitatively and qualitatively. For example, our PSNR score is 1.7dB better than H.264 on average under the bandwidth at 5Mbps.
	
	\section{Related Work}
	We review the related works by their categories.
	\subsection{Image/Video Compression}
	Transform coding using the discrete/integer cosine transform (DCT/ICT) has been widely used in image and video coding standards, such as JPEG \cite{Wallace_jpeg_1991}, MPEG-4 \cite{Li_mpeg_2001}, H.264 \cite{Wiegand_h264_2003}, and HEVC~\cite{sullivan2012overview}. The encoder divides an image/video frame into non-overlapping blocks, applies DCT/ICT to each individual block, and quantizes the coefficients. Because of the energy concentration ability of DCT/ICT, the compressed images are of good quality at moderate to high bit rates. However, real-time video stream requires very low bit-rate compression. As a result, the compressed images often suffer from blocking artifacts due to the block-wise processing. Another problem with existing coding standards is that they have been designed to be universal coders and cannot be tailored to specific domains.
	
	Machine learning-based techniques have been developed to aid these compression standards. A colorization model~\cite{Cheng_ICML_2007} is utilized to learn the representative color pixels to store for better reconstruction. Another work adopts a dictionary-based approach~\cite{Skretting_icassp_2011} to learn sparse representations for image compression. Our method is also based on learning, in which we leverage state-of-the-art deep learning models to compress the residual information. Recently, a reinforcement learning scheme is adopted to adaptively select the bit rate for video streaming for minimizing the latency~\cite{Mao_sigcomm_2017}. 
	
	\subsection{Deep Learning-based Image Compression}
	
	Instead of engineering every step in the coding system, numerous learning-based approaches as discussed in Jiang et al.~\cite{Jiang_survey_1999} have been developed to compress the data in the holistic manner. Recently, autoencoders \cite{Hinton_science_2006} have been widely used in extracting abstract representations of images through learning to reconstruct the input signals~\cite{Vincent_ICML_2008,Pathak_CVPR_2016,Tsai_CVPR_2017}. While these methods use a bottleneck layer to learn a compact representation, each dimension of the compact representation is continuous, which needs to be further quantized for compression.
	
	Various approaches are utilized in the bottleneck layer to compress the data. Quantization methods~\cite{Balle_ICLR_2017,Theis_ICLR_2017} estimate the entropy rate of the quantized data while optimizing used number of bits. In addition, an adversarial training scheme~\cite{Rippel_ICML_2017} is developed to produce sharper and visually pleasing results. On the other hand, variants of the autoencoder architecture with recurrent networks show the ability to directly learn binary representations in the bottleneck layer~\cite{Toderici_ICLR_2016,Toderici_CVPR_2017}. 
	
	While theses methods only focus on still image compression, our algorithm is designed to improve video quality in streaming applications, especially in the domain-specific scenarios such as game or street view videos. To this end, we integrate existing video compression standards that can effectively exploit temporal information and learning-based methods that can efficiently transmit binary residual representations. As a result of the integration, our system can be adaptively applied to existing video compression platforms and improve their performance.
	\begin{figure*}[t]
		\centering
		\includegraphics[width=0.85\linewidth]{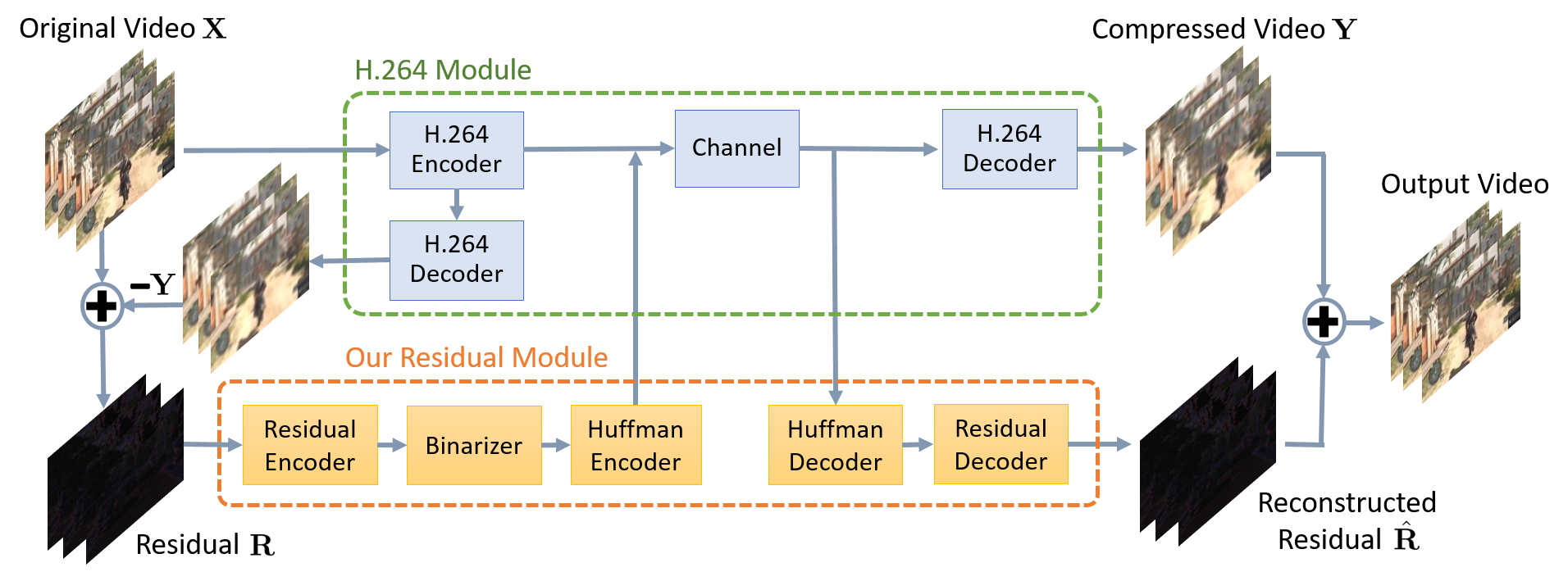}
		\caption{
			Overview of our proposed video streaming pipeline. It consists of two modules: a conventional H.264 module and our proposed residual autoencoder. The input to our residual module is the difference between the original and compressed videos. The difference is encoded and binarized to generate binary representations. We utilize Huffman coding to further compress the binary representations into a bit stream in a lossless manner. On the client side, we reconstruct the output video by adding back the decoded difference to the compressed video.}
		\label{fig:overview}
	\end{figure*}

	\subsection{Post-processing to Remove Compression Artifacts}
	Post-processing at the decoder end aims at reducing the coding artifacts without introducing additional bit rates at the encoder end. Earlier approaches~\cite{Reeve_icassp_1983,Chen_itcsvt_2001} manually use smoothing filters to reduce blocking effects caused by DCT in the expense of more blurry image outputs. Recently, learning-based methods~\cite{Chang_tip_2014,Chen_ICCV_2015,Liu_CVPR_2015,Mao_NIPS_2016} are developed to model different compression artifacts. For example, the ARCNN method~\cite{Dong_ICCV_2015} uses end-to-end training for removing various JPEG artifacts. Furthermore, the D3 scheme~\cite{Wang_CVPR_2016} employs the JPEG priors to improve the reconstruction results. Deeper models such as DDCN~\cite{Guo_ECCV_2016} are also developed to eliminate artifacts, and a recent work~\cite{Guo_CVPR_2017} combines the perceptual loss to generate visually pleasing outputs. We note that these post-processing methods require extensive computation on the client side, which is not well-suited for embedded devices. In contrast, our method encodes the residual information in a binary form on the server side and send it to the client. Utilizing the encoded residual information, the client can better recover the original video using much less computational power. 
	
	\section{Domain-specific Video Compression with Residual Autoencoder}
	
	Here, we first introduce our video compression pipeline for streaming domain-specific videos. We then present the details of our autoencoder for encoding the residual information into binary forms and the training methods. 
	
	\subsection{Video Streaming Pipeline}
	
	The proposed pipeline consists of two modules: a video compression module and a deep learning-based autoencoder, as shown in Figure~\ref{fig:overview}. The video compression module adopts the H.264 standard, which has demonstrated good performance in compressing temporally smooth visual signals, while our residual autoencoder assists to recover the lost information during compression on the client side by leveraging the property that the videos to be compressed are from the same domain. Using such a hybrid approach, not only can we improve the output quality by spending a small amount of effort, but also the system can adapt to existing compression platforms and train for specific domains by exploiting large-scale data. Note that, although we use H.264 in our pipeline, other video compression standards such as MPEG4 and HEVC can be used as well.

	Given an input video $\mathbf{X}$, we obtain the compressed video $\mathbf{Y}$ by applying H.264. The difference between the two videos is called the residual information $\mathbf{R} = \mathbf{X} - \mathbf{Y}$. The larger the residual information, the poorer the compressed video quality. We also note that $\mathbf{R}$ is not included in $\mathbf{Y}$ because it consists of highly non-linear patterns, which can not be  compressed effectively with conventional approaches.

	We argue that by limiting the video domain, we could leverage a novel autoencoder to effectively compress the residual information. The autoencoder consists of a pair of functions $(\mathcal{E},\mathcal{D})$, where the encoder $\mathcal{E}$ maps $\mathbf{R}$ to a binary map and the decoder $\mathcal{D}$ recovers $\mathbf{R}$ from the binary map on the client side. The recovered residual information is referred to as $\hat{\mathbf{R}}$ and the final output video $\mathbf{Y}+\hat{\mathbf{R}}$ has a better visual quality than $\mathbf{Y}$. We note that the binary map is further mapped to a bit stream by using the Huffman coding algorithm~\cite{Cover_information_theory}, which is asymptotically optimal, to reduce its bandwidth usage.

	Sending the bit stream of the residual information requires additional bandwidth. However, we can train an autoencoder that only requires a much smaller bandwidth to compress the residual information than H.264. Therefore, we can run the H.264 standard in a higher compression rate, which uses a smaller bandwidth but results in a larger residual signal. We then apply our autoencoder to compress the residual signal into a small bit stream. Considering a scenario where the bandwidth for a video stream is 5Mbps, we can apply the proposed pipeline to compress the video in 4Mbps using H.264 and utilize the remaining 1Mbps for sending the residual signal. Because our autoencoder is more efficient than H.264 in compressing the residual signal, our system achieve better performance than a baseline system that allocates all the 5Mbps for H.264.
	
	\begin{figure*}[t]
		\centering
		\includegraphics[width=0.85\linewidth]{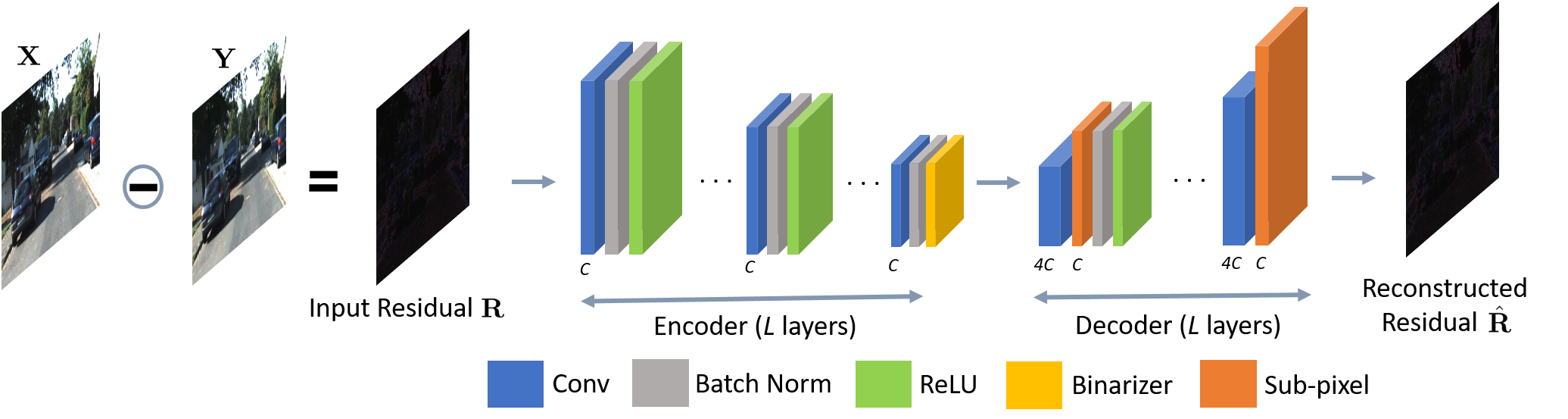}
		\caption{
			Architecture of the proposed binary residual autoencoder. It process the video frame by frame. The autoencoder consists of an encoder, a binarizer and a decoder. The encoder/decoder, each has $L$ convolutional layers and each layer contains $C$/$4C$ channels. For simplicity, Huffman coding modules are not illustrated here.
		}
		\label{fig:autoencoder}
	\end{figure*}
	
	One may wonder why not completely replacing the H.264 standard with the proposed residual autoencoder. We argue that our residual autoencoder is only more efficient than H.264 in compressing the residual signal. The carefully-engineered H.264 is more efficient in compressing the core video. By marrying the strength of H.264 and the proposed autoencoder, our hybrid system achieves better performance. Moreover, our pipeline can be easily integrated into the existing H.264 standard since the residual information can be attached in the meta field of a H.264 streaming packet. Hence we can enjoy the popularity of the H.264 standard and various hardware accelerators implemented for H.264.

	We note that in the proposed domain-specific video streaming pipeline, one needs to send the parameters of $\mathcal{D}$ and the Huffman decoder table to the client for the streaming service, which requires an additional bandwidth. However, the parameters can be sent before the streaming starts. Once the parameters are sent, the user can enjoy the low latency and high video quality features of the proposed video streaming pipeline.
	
	\subsection{Binary Residual Autoencoder}
	
	We design an autoencoder that consists of three components: encoder $\mathcal{E}$, binarizer $\mathcal{B}$ (introduced in the next section) and decoder $\mathcal{D}$. For the encoder, the goal is to learn to extract compact feature representations for the following binarizer. We use $L$ convolutional layers in our encoder, in which each layer has the same channel number $C$ and a stride of two that down-samples feature maps. The binarizer converts the output from the last convolutional layer into a binary map. For the decoder, we aim to up-sample the binary map back to the original input. Our decoder has $L$ convolutional layers. At the end of each convolutional layer, a sub-pixel layer~\cite{Shi_CVPR_2016} is used for up-sampling. Since we aim for upsampling the resolution of the feature map by two on each spatial dimension, the channel number of each convolutional layer in the decoder is set to $4\times C$ due to the use of the sub-pixel layer. To facilitate the learning process, batch normalization~\cite{Ioffe_ICML_2015} and ReLU layers are used. The architecture of the autoencoder are given in Figure~\ref{fig:autoencoder}.

	We encode and decode the residual signal $\mathbf{R}$ frame by frame. Let $\{r_i\}$ be a set of residual frames computed by applying H.264 at a target bit rate. We train our autoencoder by solving the following optimization problem:
	\begin{equation}
	\min_{\mathcal{D},\mathcal{E}} \sum_{i}||r_i -\mathcal{D}(\mathcal{B}(\mathcal{E}(r_i)))||_2^2.
	\label{eq:l2}
	\end{equation}
	We care a lot about the bandwidth required for transmitting binary representations, which is determined by two factors: 1 the number of layers $L$ in the encoder, and 2) the number of channels $C$. Let $W$ and $H$ be the width and height of the input image, the binary map size is given by $\frac{C\times W\times H}{2^{2L}}$. A large number of $L$ and a smaller number of $C$ would result in a smaller size of the encoder output and hence a smaller binary map. However, a smaller encoder output makes training of the autoencoder difficult. In our experiments we will discuss the results with different numbers of $L$ and $C$.
	
	\subsection{Training of the Binary Residual Autoencoder}
	
	Binarizing feature maps in neural networks have been studied in several earlier works \cite{Rastegari_ECCV_2016,Courbariaux_corr_2016,Tang_AAAI_2017} for the purpose of reducing memory footprint in mobile devices. It is also used for image compression \cite{Toderici_ICLR_2016,Toderici_CVPR_2017}, while our work is different in that we binarize feature maps for video streaming. We will discuss several binarization methods here and compare their advantages and drawbacks when used in our pipeline in the experiment section.
	
	{\flushleft {\bf Formulation.}}~Let the output feature map of $\mathcal{E}$ be $e_i = \mathcal{E}(r_i)$. Our binarizer $\mathcal{B}$ aims to produce the binary output $\{-1, 1\}$\footnote{We have found that our network requires negative responses to achieve reasonable results. Instead of producing the binary output $\{0, 1\}$, we aim to generate the discrete output $\{-1, 1\}$.}. To generate such binary outputs, the process $\mathcal{B}$ consists of two parts: 1) map each element of the encoder output $e_i$ to the interval $[-1, 1]$, and 2) discretize it to $\{-1, 1\}$ given by:
	\begin{equation}
	\mathcal{B}(e_i) = b(\sigma(e_i)),
	\label{eq: binary}
	\end{equation}
	where $\sigma$ and $b$ are the activation and discretization functions, respectively. In the following, we discuss different functions for activation (i.e., \textit{tanh}, \textit{hardtanh}, \textit{sigmoid}) and various methodologies for binarization (i.e., stochastic regularization \cite{Raiko_ICLR_2015} , Gumbel noise \cite{Jang_ICLR_2017,Maddison_ICLR_2017}).
	
	{\flushleft {\bf Tanh/Hardtanh Activation.}} \textit{tanh} is a common activation to project feature values to $[-1, 1]$, so as the approximation version \textit{hardtanh} function. Here, we define the binarized output as:
	\begin{align}
	b(z) =
	\begin{cases}
	1, &  \text{if $z\geq 0$} \\
	-1, &  \text{if $z< 0,$}
	\end{cases}
	\label{eq: thres}
	\end{align}
	where $z = \sigma(e_i)$ and $\sigma$ can be the \textit{tanh} or \textit{hardtanh} function. However, since binarization is not a differentiable function, we can not train the proposed autoencoder using back-propagation. To avoid the issue, inspired by the recent binarization work~\cite{Courbariaux_corr_2016}, we adopt a piecewise function $b_{bp}$ during back-propagation:
	\begin{align}
	b_{bp}(z) =
	\begin{cases}
	1, &  \text{if $z> 1$} \\
	z, & \text{if $-1 \leq z \leq 1$} \\
	-1, &  \text{if $z< -1.$}
	\end{cases}
	\label{eq: bp}
	\end{align}
	By using the straight-through estimator \cite{Courbariaux_corr_2016}, we can compute the gradient of $b_{bp}(z)$ and pass gradients through $b$ unchanged as:
	\begin{align}
	b_{bp}'(z) =
	\begin{cases}
	1, & \text{if $-1 \leq z \leq 1$} \\
	0, &  \text{otherwise.}
	\end{cases}
	\label{eq: gradient}
	\end{align}

	{\flushleft {\bf Sigmoid Activation.}}
	The \textit{sigmoid} function outputs a value in $[0, 1]$. We convert the output to $[-1, 1]$ by applying $z = 2(\sigma(e_i)-0.5)$. We can then use the approach discussed in the previous paragraph for binarization and training.
	
	{\flushleft {\bf Stochastic Regularization.}} Following Toderici et al.~\cite{Toderici_ICLR_2016}, we incorporate a stochastic process into \eqref{eq: thres} using the \textit{tanh} activation:
	\begin{equation}
	b(z) = z + \epsilon, \notag
	\end{equation}
	where $\epsilon$ is a randomized quantization noise, resulting in:
	\begin{align}
	b(z) =
	\begin{cases}
	1, & \text{with probability $\frac{1+z}{2}$} \\
	- 1, &  \text{with probability $\frac{1-z}{2}.$}
	\end{cases}
	\label{eq: stochastic}
	\end{align}
	For back-propagation, similarly we can pass unchanged gradients through $b$ \cite{Toderici_ICLR_2016}.
	
	{\flushleft {\bf Gumbel Noise.}} The Gumbel-Softmax distributions have been utilized for learning categorical/discrete outputs \cite{Jang_ICLR_2017}. We adopt a similar approach by adding Gumbel noise in the \textit{sigmoid} function, which we refer as Gumbel-Sigmoid. Since \textit{sigmoid} can be viewed as a special 2-class case ($e_i$ and 0 in our case) of \textit{softmax}, we derive the Gumbel-Sigmoid as:
	\begin{equation}
	\sigma(e_i) = \frac{\exp((e_i + g_k)/\tau)}{\exp((e_i + g_k)/\tau) + \exp(g_l/\tau)},
	\label{eq: gumbel}
	\end{equation}
	where $g$ is the sample from the Gumbel noise and $\tau$ is the temperature that controls how closely the Gumbel-Sigmoid distribution approximates the binary distribution. From \eqref{eq: gumbel}, we can further simplify it as:
	\begin{align}
	\sigma(e_i) & = \frac{1}{1 + \exp(-(e_i + g_k - g_l)/\tau)} \notag \\ 
	& = sigm((e_i + g_k - g_l)/\tau),
	\label{eq: gumbel_sigmoid}
	\end{align}
	where $sigm$ is the \textit{sigmoid} function. Since $\sigma(e_i)$ is still in the range of $[0, 1]$, we adopt the same approach as introduced before to shift the value via $z = 2(\sigma(e_i)-0.5)$. Following \cite{Jang_ICLR_2017}, we start with a high temperature and gradually anneal it to a small but non-zero temperature.
	
	\begin{table}[t]
		\caption{Compression ratio versus number of bits in a group computed from our KITTI dataset experiments.
		}
		\small
		\centering   
		%		\begin{tabular}{|c|c|c|c|c|}
		\begin{tabular}{ccccc}
			%\hline
			\toprule
			\# of bits in a group & 8 & 16 & 32 & 64 \\
			\hline			
			Compression ratio & 1.05 & 1.47 & 3.03 & 6.67 \\
			%			\hline
			\bottomrule
		\end{tabular}
		\label{tab:entropy}
	\end{table}
	
	{\flushleft {\bf Lossless Compression.}} Once we generate the binary feature map, we further use Huffman coding \cite{Cover_information_theory} to reduce the size of the binary representation. These coding schemes are asymptotically optimal, which means that the more the number of bits we group together as a symbol, the better the compression ratio. Such a behavior is illustrated in Table \ref{tab:entropy} using the KITTI dataset. We note that other source coding methods such as arithmetic coding \cite{Marpe_itcsvt_2003} can be used as well.

	\section{Experimental Results}
	
	We evaluate our pipeline using the KITTI~\cite{Geiger_CVPR_2012} dataset, which consists of various driving sequences of street scenes, and three popular video games: Assassins Creed, Skyrim and Borderlands. The details of the datasets are shown in Table \ref{tab:dataset}. We use the tracking benchmark on the KITTI dataset that contains 50 street-view videos. We randomly select 42 videos for training and 8 videos for testing. The images in the videos are resized to $360 \times 1200$. The game video resolutions are $720 \times 1280$.
	
	We report PSNR and SSIM scores at different bit rates for quantitative comparisons. These two metrics are popular performance metrics for benchmarking video compression algorithms~\cite{Wang_TIP_2004}. We first conduct an ablation study using the KITTI dataset where we discuss the impact of layer and channel numbers in our design. We then compare various methods for training the autoencoder. Finally, we report the performance with comparisons to H.264 and a deep learning baseline.
	\begin{table}[t]
		\caption{Number of videos and frames on the datasets.
		}
		\small
		\centering   
		%		\begin{tabular}{|c|c|c|c|c|}
		\begin{tabular}{ccccc}
			\toprule
			& KITTI & Assassins Creed & Skyrim & Borderlands \\
			\hline
			
			Videos & 50 & 50 & 9 & 19 \\
			\hline
			
			Frames & 19,057 & 34,448 & 9,337 & 8,752\\		
			\bottomrule
		\end{tabular}
		\label{tab:dataset}
	\end{table}
	%
	% 	\subsection{Implementation Details}
	{\flushleft {\bf Implementation Details.}} Throughout the paper, we use Adam~\cite{Kingma_ICLR_2015} to train our binary residual autoencoder. The learning rate is set to $10^{-3}$ and then decreased by half for every 5 epochs. The momentums are set to $0.9$ and $0.999$. The batch size is $10$, and we train the model for $50$ epochs. The implementation is based on PyTorch.
	
	{\flushleft {\bf Runtime Analysis.}} On the server side, our encoder and the binarizer takes about 0.001 seconds to compress the residual image with a resolution of $360 \times 1200$ using a Titan X GPU. The decoder on the client side takes 0.001 seconds to reconstruct the residual image. %Compared to the artifact-removal model that takes twice of the time on the client side, our method is more efficient and effective in generating better results.
	%
	%	\begin{itemize}
	%		\item datasets: three games, KITTI (TODO)
	%		\item evaluation metric - bitrates v.s PSNR, SSIM
	%		\item implementation details
	%	\end{itemize}
	%
	%\color{blue}
	%
	
	\subsection{Ablation Study}
	
	{\flushleft {\bf Depth and Breadth of the Model.}} We analyze the impact of layer numbers $L$ and channel numbers $C$ on the video compression performance of our pipeline. For this study, the bandwidth of H.264 is set to 5Mbps for generating training and testing videos. We then train the residual autoencoder with different $L$ and $C$ utilizing the \textit{hardtanh} activation for binarization. These two parameters determine the size of the binary map, which has impacts on the compression rate as well as the training difficulty. Intuitively, the smaller the binary map, the easier the compression task but the harder the training task. The results are shown in Table~\ref{tab:bitrate}. Based on the bit rate, we divide the settings into three groups. The ones in the same group have a very similar bit rate after applying the Huffman coding. We then use the best setting in each bit rate group for the rest of the experiments.
	
	\begin{table}[t]
		\caption{Impact of $L$ and $C$ on PSNR/SSIM.}
		\centering   
		%		\begin{tabular}{|c|c|c|c|}
		\begin{tabular}{ccccc}
			\toprule
			Group & Mbps & ($C$, $L$) & PSNR & SSIM \\
			\hline
			\multirow{2}{*}{1} & \multirow{2}{*}{$\sim$0.24} &$(8, 3)$ & 29.28 & 0.8571 \\
			
			& & $(32, 4)$ & \textbf{29.39} & \textbf{0.8577} \\
			\hline
			
			\multirow{2}{*}{2} & \multirow{2}{*}{$\sim$0.48} & $(16, 3)$ & \textbf{29.83} & \textbf{0.8681} \\
			
			& & $(64, 4)$ & 29.81 & 0.8652 \\
			\hline
			
			\multirow{2}{*}{3} & \multirow{2}{*}{$\sim$0.96} & $(8, 2)$ & 30.35 & 0.8855 \\
			
			& & $(32, 3)$ & \textbf{30.67} & \textbf{0.8901} \\
			\bottomrule
		\end{tabular}
		\label{tab:bitrate}
	\end{table}
	%
	%	\begin{itemize}
	%		\item use the tracking dataset
	%		\item contain 47 sequences, 18062 images
	%		\item resize to 360 $\times$ 1200
	%		\item still test on training set?
	%		\item train models with different bitrate combinations
	%		\item compare different binarization methods?	
	%	\end{itemize}
	%
	\begin{figure*}[t]
		\centering
		\begin{tabular}
			{
				@{\hspace{0mm}}c@{\hspace{0mm}} @{\hspace{0mm}}c@{\hspace{0mm}} @{\hspace{0mm}}c@{\hspace{0mm}} @{\hspace{0mm}}c@{\hspace{0mm}}
			}
			\includegraphics[width=0.25\linewidth]{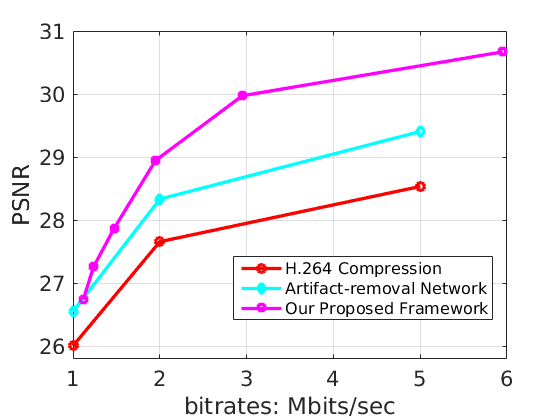} &
			\includegraphics[width=0.25\linewidth]{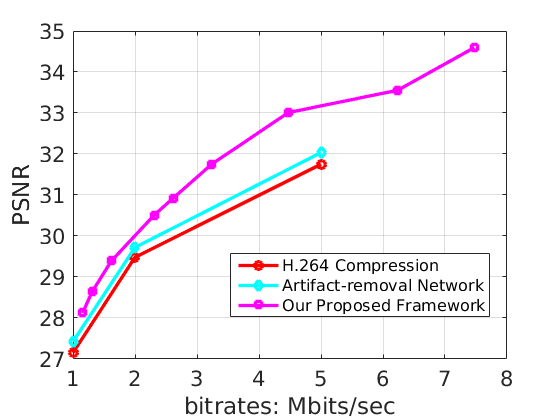} &
			\includegraphics[width=0.25\linewidth]{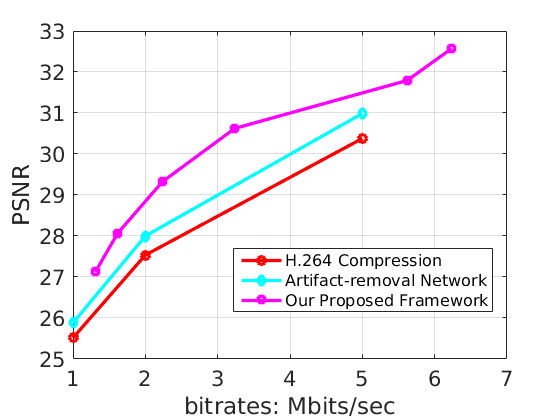} &
			\includegraphics[width=0.25\linewidth]{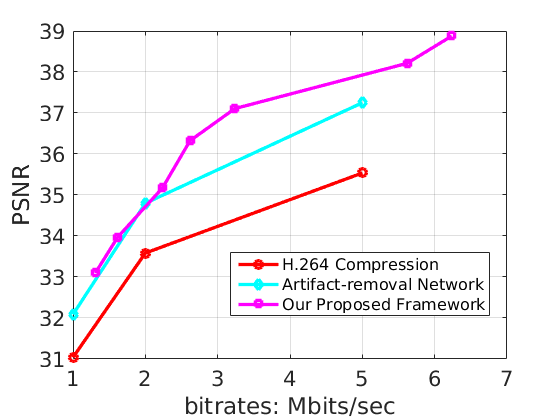} \\
			
			(a) KITTI & (b) Assassins Creed & (c) Skyrim & (d) Borderlands \\
			
		\end{tabular}
		\caption{PSNR comparisons on four datasets at different bandwidths. 
			We compare our pipeline with H.264 and an artifact-removal method based on~\cite{Kim_CVPR_2016,Zhang_TIP_2017}.
		}
		\label{fig:psnr}
	\end{figure*}
	\begin{figure*}[t]
		\centering
		\begin{tabular}
			{
				@{\hspace{0mm}}c@{\hspace{0mm}} @{\hspace{0mm}}c@{\hspace{0mm}} @{\hspace{0mm}}c@{\hspace{0mm}} @{\hspace{0mm}}c@{\hspace{0mm}}
			}
			
			\includegraphics[width=0.25\linewidth]{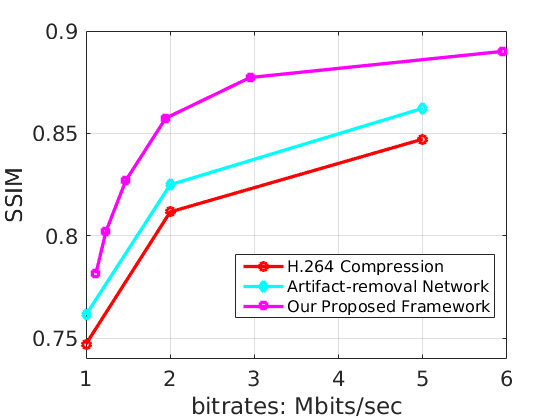} &
			\includegraphics[width=0.25\linewidth]{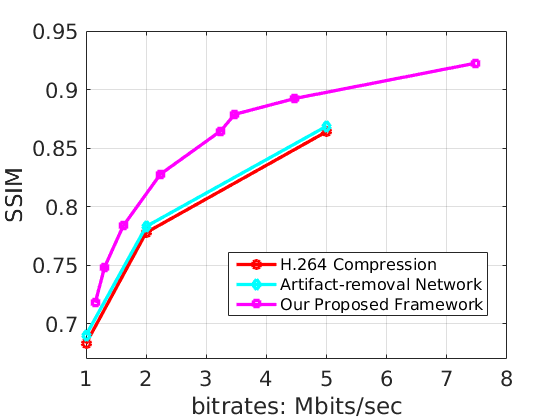} &
			\includegraphics[width=0.25\linewidth]{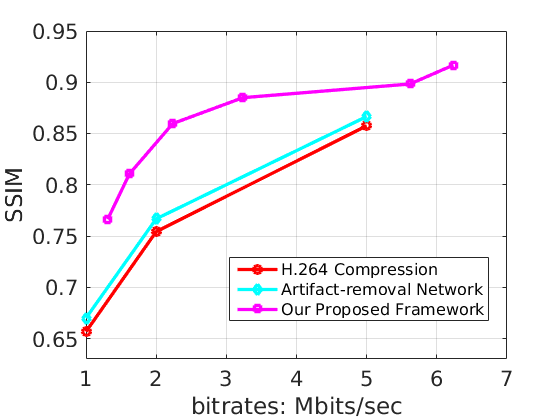} &
			\includegraphics[width=0.25\linewidth]{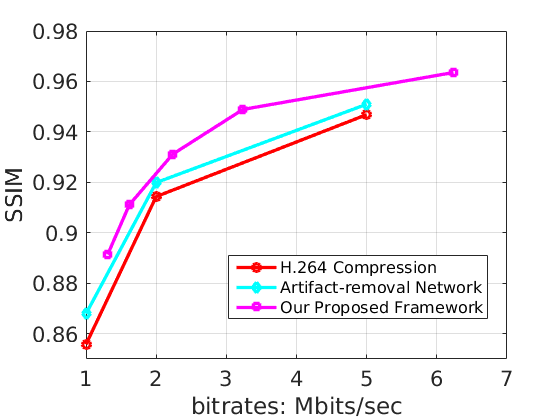} \\
			
			(a) KITTI & (b) Assassins Creed & (c) Skyrim & (d) Borderlands \\
			
		\end{tabular}
		\caption{SSIM comparisons on four datasets at different bandwidths. We compare our pipeline with H.264 and an artifact-removal method based on~\cite{Kim_CVPR_2016,Zhang_TIP_2017}.
		}
		\label{fig:ssim}
	\end{figure*}
	\begin{figure*}[!th]
		\centering
		\includegraphics[width=1\linewidth]{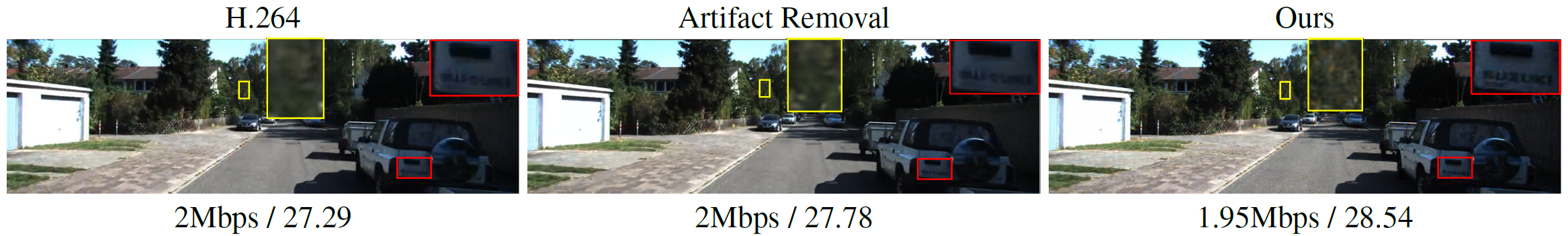} \\
		
		\includegraphics[width=1\linewidth]{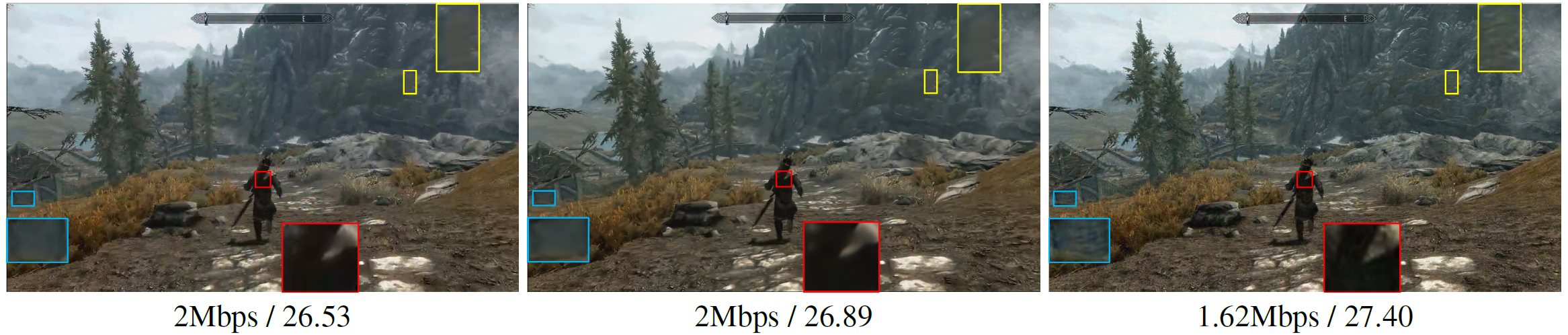} \\
		
		\includegraphics[width=1\linewidth]{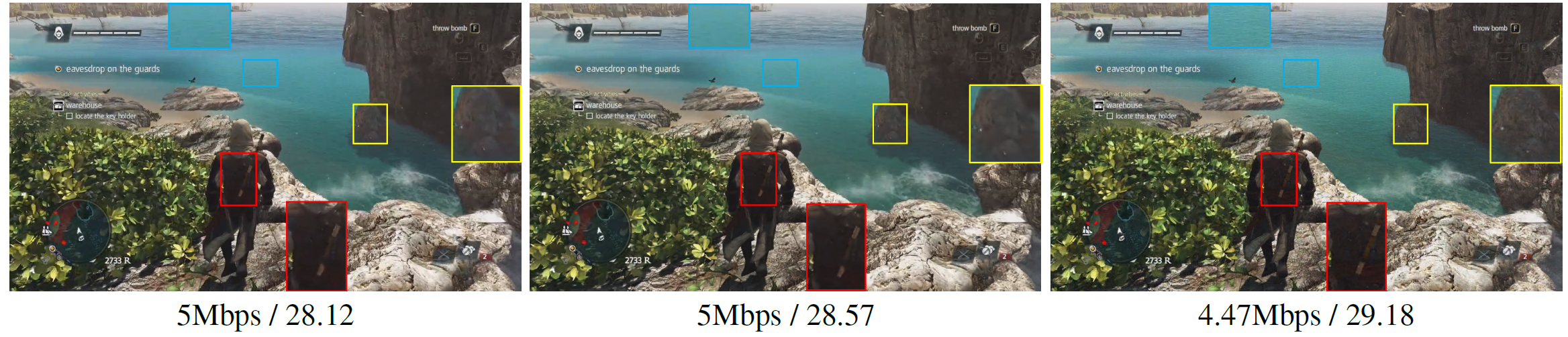} \\
		
		\caption{Example results on the KITTI and video game datasets. We compare our pipeline with H.264 and an artifact-removal method. The corresponding bit rate and PSNR are shown next to the images. Best viewed with enlarged images.
		}
		\label{fig:result}
	\end{figure*}    
	
	{\flushleft {\bf Binarization.}} We compare various methods discussed earlier for training the binary residual autoencoder. We fix $L=3$ and $C=32$ and train our model with different H.264 compression rates. The results are reported in Table \ref{tab:binary_psnr} and \ref{tab:binary_ssim}. We find that the models trained using \textit{hardtanh} and \textit{tanh} activations consistently outperform the others. The model trained with the stochastic regularization does not perform well possibly due to the training difficulty. It is known that the stochastic noise increases the gradient variance. In addition, we empirically find that the model using Gumbel-Sigmoid performs much worse for the residual signal compression task. (We note that \cite{Jang_ICLR_2017} uses the Gumbel noise for reconstructing discrete/categorical signals but not continuous signals of images in our task.). Hence, we use \textit{hardtanh} activations in the rest of the experiments for its superior performance.

	\begin{table}[t]
		\caption{Comparisons of different binarization methods on the KITTI dataset using PSNR.
		}
		\centering   
		%		\begin{tabular}{|c|c||c||c|}
		\begin{tabular}{cccc}
			\toprule
			H.264 $@$ Mbps & 1M & 2M & 5M \\
			%			\hline
			%			\hline
			%			H.264 & 26.01 & 27.66 & 28.54 \\
			\hline
			
			Hardtanh & \textbf{28.95} & \textbf{29.97} & \textbf{30.67} \\
			\hline
			
			Tanh & \textbf{28.95} & \textbf{29.97} & \textbf{30.67} \\
			\hline
			
			Sigmoid & 28.89 & 29.88 & 30.61 \\
			\hline
			
			Stochastic & 28.26 & 29.21 & 29.78 \\
			\hline
			
			Gumbel-Sigmoid & 26.66 & 27.82 & 28.66 \\
			\bottomrule
		\end{tabular}
		\label{tab:binary_psnr}
	\end{table}
	
	\subsection{Video Compression Performance}
	
	We compare our video compression pipeline to the H.264 standard and an artifact removal network, which is a popular approach to reduce distortions. Following recent deep learning works in image enhancement and denoising \cite{Kim_CVPR_2016,Zhang_TIP_2017}, we utilize a neural network with 8 convolutional layers (no strides) to remove the artifacts. The network takes the compressed H.264 images as inputs in a frame-by-frame fashion and outputs enhanced images. We then train an artifact removal network for each video domain for fair comparisons.
	
	We note that the proposed pipeline requires bandwidth for both H.264 and the binary map. We account both in the bit rate calculation for fair comparisons with the baseline methods. In other words, the bit rate of the proposed pipeline is the sum of the bit rate from the H.264 stream and the Huffman code of the binary map. Again, we do not take transmitting the network parameters into account since it can be sent before the streaming starts. In the streaming services, the main goal is to have high quality videos with low latency.
	
	We report PSNR and SSIM scores on the KITTI benchmark and 3 video games in Figures \ref{fig:psnr} and \ref{fig:ssim}. We find our pipeline consistently outperforms the baseline methods at all bit rates. Our pipeline achieves a PSNR of $33.26$dB at 5Mbps averaged on four datasets, which is $1.71$dB better than H.264 and $0.84$dB better than the artifact-removal network. Similarly, our pipeline performs better in SSIM, e.g., $5.3\%$ and $4.1\%$ improvements over H.264 and the artifact-removal network at 2Mbps, respectively. In Figure \ref{fig:result}, we present some qualitative results, showing our method preserves more details and textured contents (e.g., tree, rock and water) in reconstructed images using a smaller bandwidth\footnote{The website link http://research.nvidia.com/publication/2018-02\_Learning\-Binary\-Residual contains more visualization results.}.

	To validate the importance of modeling the residual, we also carry out experiments by directly compressing the raw video using the same autoencoder architecture. However, it results in a worse performance compared to H.264 (0.52dB drop in PSNR) since this method does not leverage any motion information for video compression. 	Overall, our results show that by propagating the extracted binary residual representations from the server to the client, the quality of reconstructed videos can be largely improved. It outperforms the artifact removal network, which aims for solving a challenging inverse problem and does not leverage any prior knowledge about the compression process. In addition, the runtime of our binary residual autoencoder (decoding on the client side) is two times faster than the artifact removal network.
	
	\begin{table}[t]
		\caption{Comparisons of different binarization methods on the KITTI dataset using SSIM.
		}
		\centering   
		%		\begin{tabular}{|c|c||c||c|}
		\begin{tabular}{cccc}
			\toprule
			H.264 $@$ Mbps & 1M & 2M & 5M \\
			%			\hline
			%			\hline
			%			H.264 & 0.7473 & 0.8118 & 0.8473 \\
			\hline
			
			Hardtanh & 0.8575 & \textbf{0.8773} & \textbf{0.8901} \\
			\hline
			
			Tanh & \textbf{0.8577} & 0.8770 & 0.8882 \\
			\hline
			
			Sigmoid & 0.8538 & 0.8722 & 0.8861 \\
			\hline
			
			Stochastic & 0.8297 & 0.8470 & 0.8617 \\
			\hline
			
			Gumbel-Sigmoid & 0.7683 & 0.8128 & 0.8464 \\
			\bottomrule
		\end{tabular}
		\label{tab:binary_ssim}
	\end{table}

	\section{Conclusions}
	In this paper, we propose a video streaming system that integrates H.264 and a binary residual autoencoder to encode non-linear compression errors for domain-specific video streaming. We analyze various network design choices and methods for obtaining binary representations of the residual information. The binary representations are further compressed and transmitted from the server to the client. On the KITTI benchmark dataset and three popular video game datasets, the proposed algorithm generates better reconstructed videos than H.264 and artifact-removal methods while using a smaller bandwidth.

	{\bf \noindent Acknowledgment} The authors would like to thank Sam H. Azar and Zheng Yuan for their helpful discussions.
	
	\bibliography{mybib}
	\bibliographystyle{aaai}
\end{document}